\pdfoutput=1 
\documentclass[11pt,a4paper]{article}
\usepackage[]{emnlp2021}
\usepackage{times}
\usepackage{latexsym}
\usepackage{booktabs}
\usepackage{makecell}
\usepackage{multirow}
\usepackage{threeparttable}
\usepackage{paralist}
\usepackage{placeins}
\usepackage{float}
\usepackage{amsmath}
\usepackage{amssymb}
\usepackage{stfloats}

\usepackage{graphicx}
\usepackage{lettrine}
\newcommand{\GrantNo}{INEA/CEF/ICT/A2019/1927024}
\newcommand{\ProjectName}{User-Focused Marian}
\newcommand{\ProjectType}{Connecting Europe Facility}

\usepackage{microtype}

\newcommand\blfootnote[1]{%
  \begingroup
  \renewcommand\thefootnote{}\footnote{#1}%
  \addtocounter{footnote}{-1}%
  \endgroup
}

\title{The Highs and Lows of Simple Lexical Domain Adaptation Approaches for Neural Machine Translation}

\author{Nikolay Bogoychev\textsuperscript{*} \qquad\qquad Pinzhen Chen\textsuperscript{*} \\
  School of Informatics, University of Edinburgh \\
  \texttt{\{n.bogoych, pinzhen.chen\}@ed.ac.uk}}

\begin{document}
\maketitle

\begin{abstract}

Machine translation systems are vulnerable to domain mismatch, especially in a low-resource scenario. Out-of-domain translations are often of poor quality and prone to hallucinations, due to exposure bias and the decoder acting as a language model. 
We adopt two approaches to alleviate this problem: lexical shortlisting restricted by IBM statistical alignments, and hypothesis re-ranking based on similarity. The methods are computationally cheap, widely known, but not extensively experimented on domain adaptation. We demonstrate success on low-resource out-of-domain test sets, however, the methods are ineffective when there is sufficient data or too great domain mismatch. This is due to both the IBM model losing its advantage over the implicitly learned neural alignment, and issues with subword segmentation of out-of-domain words. 
\end{abstract}

\section{Introduction}

\blfootnote{* Equal contribution.}
Neural Machine translation (NMT) has achieved state-of-the-art performance in a variety of language pairs and settings \citep{bahdanau_nmt, vaswani_transformer}, but it is vulnerable to domain mismatch, where the test set differs significantly from the training data in terms of vocabulary, genre, length, etc. This issue is exacerbated in a low-resource condition \citep{koehn-knowles-2017-six}.

Teacher forcing is used during traditional maximum likelihood neural network training, leading to a strong exposure bias, and model confusion when presented with unexpected sequences. This typically results in hallucinations in the output \citep{mller2019domain}, because the overly zealous language model component prefers a fluent translation, as opposed to an adequate one. A number of methods have been proposed in order to tackle the issue: exposing the model to its predictions during training \citep{ranzato2015sequence, shen-etal-2016-minimum, zhang-etal-2019-bridging, rico-exposure}; tuning directly towards BLEU \citep{wiseman-rush-2016-sequence} or using minimum Bayes risk decoding \citep{kumar-byrne-2004-minimum, stahlberg-etal-2017-neural}.
A common weakness of such methods is that they are computationally expensive.

In this paper, we adopt and experiment with two approaches inspired by previous research. We use lexical shortlisting to interpolate a statistical alignment model with NMT; on top of it, we perform n-best list re-ranking by hypothesis agreement. Our aim is to constrain the lexical choice of the decoder, to prevent hallucinations from being generated. The methods are computationally simpler, as they require no change to the model or training. We analyse the effectiveness of these methods in different scenarios of domain adaptation. We show BLEU gains on a variety of out-of-domain datasets in a low-resource English-German setting. However, the methods show no improvements once the datasets are large, or the domains are too distant. 

\section{Methodology}
\label{sec:method}
We purse and analyse two separate strategies for improving neural machine translation's performance in low-resource domain mismatched settings: lexical shortlisting and n-best list re-ranking based on inter hypothesis agreement.

\subsection{Lexical shortlisting}
Neural machine translation systems have a vocabulary size of tens of thousands, but for every single translation, most of the vocabulary items are improbable choices. Many researchers attempt to improve on this to speed up the computation in the output layer \citep{schwenk-etal-2007-smooth,le-etal-2012-continuous, devlin-etal-2014-fast}. This is done by preparing a list of likely word-level translations for each sentence (commonly known as lexical shortlist), by using IBM alignment models such as fast-align \citep{dyer-etal-2013-simple} and limiting the output layer choices to it. This has been implemented in NMT frameworks for efficient systems \citep{junczys-dowmunt-etal-2018-marian-cost} and since widely used \citep[][inter alia]{bogoychev-etal-2020-edinburghs}.

Recently, \citet{li-etal-2019-word} showed that even in a high-resource scenario, the quality of the IBM alignments can outperform those learned by a neural model. Therefore, we find it sensible to incorporate word alignments into neural models, especially in a low-resource situation. 
Although shortlists have a negligible impact on the quality of strong neural systems, we can successfully limit the output layer to the likely tokens according to an IBM model trained on a particular domain, and improve out-of-domain BLEU scores in a low-resource setting. This suggests that the IBM model provides complementary information to the neural model in this scenario.

\subsection{Hypothesis re-ranking}
\citet{fomicheva-2020-multi-hypothesis-eval} estimate the quality of machine translation by measuring the agreement of the generated hypotheses for a given source sentence. The reason is that (higher) similarity between hypotheses reflects (higher) model confidence. In our problem, we assume that a hallucinated hypothesis will have a low agreement with the rest in the beam. Thus, we propose to re-rank the n-best list based on inter-hypothesis similarity and select the top one as the final translation. 

For every source sentence, NMT generates $b$ hypotheses where $b$ is the beam size. For each hypothesis, we measure its similarity against others in the beam, then sort all hypotheses by their final aggregated similarity scores. The similarity score $score_i$ for a certain hypothesis $hyp_i$ is calculated as Equation~\ref{eq-similarity}, where the similarity between $hyp_i$ and every other $hyp_j$ produced by the model is measured by an automatic metric $similarity$:
\begin{equation}
    \text{score}_i = \sum_{j=0, i \neq j}^{b}\text{similarity}(\text{hyp}_i, \text{hyp}_j)
    \label{eq-similarity}
\end{equation}

\section{Experimental setup}

For our experiments, we use OPUS English-German data \citep{opus-lison-tiedemann-2016-opensubtitles2016}, with preprocessing performed as per the work of \citet{mller2019domain}. 
We follow the same data split as \citet{rico-exposure}'s low-resource, domain restricted setting: a model is trained on 1M sentences pairs from the \textit{medical} domain, and is evaluated on \textit{medical} (in-domain), \textit{Koran}, \textit{IT}, \textit{subtitles} and \textit{law} (out-of-domain). We use a joint vocabulary byte-pair encoding \citep[BPE,][]{subword_nmt} trained on the \textit{medical} domain with 32k merge operations. In this way, the vocabulary trained on the \textit{medical} domain may be sub-optimal for the out-of-domain test sets. In reality, monolingual data from unknown domains might be available, which can reduce the bias of the BPE training towards the in-domain data. In order to simulate this scenario, we perform an analogous experiment where the BPE vocabulary is trained on all domains, except the \textit{subtitles} corpus, as it is much larger than other domains and would dominate the vocabulary (reasons elaborated in Section~\ref{sec:limitations}).

For training, we used the Transformer-base preset of the Marian toolkit \citep{2018-marian} with transformer preprocessing normalisation as opposed to postprocessing normalisation, and additional attention and feed-forward layer dropout.

We performed ample hyperparameter search, to ensure that our neural network configuration achieves the best possible performance on this low-resource task. Decoding is always done with beam size 6 and length normalisation 0.6. On top of the baseline, we tried three different combinations of methods introduced in Section~\ref{sec:method}: 
\begin{itemize}
    \item A \textbf{shortlisting} configuration where we use a lexical shortlist generated by the \textit{fast-align} model trained on the \textit{medical} dataset. We try various configurations, and we settle on an optimal value of limiting the output layer to the 10 most probable unigram translations according to the IBM model.
    
    \item A \textbf{re-ranking} setup where we re-rank the n-best translation based on inter-hypothesis similarity. Like \citeauthor{fomicheva-2020-multi-hypothesis-eval}, we tested out a few different metrics to measure similarity: sentBLEU, ChrF, TER and METEOR. In an initial experiment on the \textit{medical} domain, we found sentBLEU to have the best performance and stick to it for all experiments. We did not pick any neural metrics, as they may have a domain preference and are less interpretable.
    
    \item \textbf{Both}: shortlisting followed by re-ranking as specified above
\end{itemize}

\section{Results and Analysis}
\label{sec_results_and_analysis}

\begin{table*}[!ht]
\centering
\begin{tabular}{|l|cccc|cccc|}
\hline
\multirow{2}{*}{Domain} & \multicolumn{4}{c|}{BPE trained on \textit{medical} only}                       & \multicolumn{4}{c|}{BPE trained on all except \textit{subtitles}}               \\ \cline{2-9} 
          & baseline & shortlist     & re-rank & both         & baseline & shortlist     & re-rank & both          \\ \hline
medical & \textit{60.0} & \textit{59.5} & \textit{\textbf{60.3}} & \textit{59.1} & \textit{\textbf{61.4}} & \textit{58.2} & \textit{57.6} & \textit{60.4} \\
Koran     & 0.9      & 1.0           & 0.7     & \textbf{1.1} & 0.8      & 0.9           & 0.9     & \textbf{1.0}  \\
law       & 19.6     & \textbf{20.6} & 16.6    & 17.8         & 17.8     & 19.3          & 19.8    & \textbf{20.8} \\
IT        & 15.0     & \textbf{16.3} & 10.1    & 11.5         & 15.7     & \textbf{18.0} & 15.3    & 17.8          \\
subtitles & 2.8      & \textbf{3.1}  & 1.4     & 1.9          & 2.6      & \textbf{2.8}  & 2.4     & \textbf{2.8}  \\ \hline
\end{tabular}

\caption{BLEU results on German-English systems trained on the \textit{medical} domain, and tested on in-domain and out-domain datasets. In-domain results are in italics and the best BLEU on each domain dataset are in bold.}
\label{results_tab}
\end{table*}

\begin{table*}[tbhp]
    \centering
    \begin{tabular}{|c|c|c|c|c|c|c|c|c|}
    \hline
     Domain & System & \multicolumn{4}{c|}{1- to 4-gram precisions} & \makecell{ Brevity \\penalty} &  \ BLEU \ ($\triangle$) & \ METEOR \ ($\triangle$) \\
\hline
\multirow{4}{*}{law} & baseline  & 53.0 & 27.5 & 16.9 & 11.0 & 0.778 & 17.8 \hspace{30pt} &  0.36 \hspace{35pt} \\
         & shortlist &  \textbf{56.1} & \textbf{29.4} & \textbf{17.9} & \textbf{11.4} & 0.804 & 19.3 ($+$1.5) & \textbf{0.39 ($+$0.03)} \\
         & re-rank & 51.4 & 26.4 & 16.1 & 10.5 & 0.906 & 19.8 ($+$2.0) & 0.31 ($-$0.05) \\
         & both & 53.1 & 27.6 & 16.7 & 10.7 & \textbf{0.919} & \textbf{20.8 ($+$3.0)} & 0.35 ($-$0.01) \\\hline
         
\multirow{4}{*}{IT} & baseline & 34.6 & 18.8 & 13.1 & 9.5 & 0.930 & 15.7 \hspace{30pt} & 0.16 \hspace{35pt} \\
         & shortlist & \textbf{43.9} & \textbf{24.7} & \textbf{17.1} & \textbf{12.1} & 0.828 & \textbf{18.0 ($+$2.3)} & \textbf{0.18 ($+$0.02)} \\
         & re-rank & 33.5 & 17.2 & 11.7 & 8.1 & \textbf{1.000} & 15.3 ($-$0.4) & 0.12 ($-$0.04) \\
         & both & 38.0 & 20.1 & 13.6 & 9.7 & \textbf{1.000} & 17.8 ($+$2.1) & 0.09 ($-$0.07) \\\hline
    \end{tabular}
    \caption{Breakdown of BLEU scores for ``BPE on all'' experiments in Table~\ref{results_tab} for \textit{law} and \textit{IT} domains.}
    \label{table-bleu-scores}
\end{table*}

We present in Table~\ref{results_tab} our models' BLEU scores on in- and out-of-domain test sets, with both BPE segmentation schemes. When applying the more restrictive BPE trained on the \textit{medical} domain, our shortlisting mechanism always yields a slight increase in BLEU on out-of-domain sets and a small drop on the in-domain set. The \textit{law} and \textit{IT} domains benefit the most, whereas \textit{subtitles} and \textit{Koran} are largely unaffected. Re-ranking is not helpful in this BPE setting. 

When using the alternative BPE segmentation trained on all datasets, we see that the baseline scores are generally lower for the out-of-domain datasets, despite a stronger in-domain BLEU. This is potentially because the rare words from out-of-domain datasets get insufficient exposure during training (on \textit{medical} only). However, when using a lexical shortlist in this setting, we see greater improvements in terms of BLEU on the \textit{law} and \textit{IT} domains compared to the \textit{medical}-only BPE scenario. Re-ranking is also much more effective, performing similarly to shortlisting. The combination of re-ranking and shortlisting delivers the best BLEU scores in nearly all out-of-domain splits.

We see that the shortlisting method is always superior to the baseline method on all out-of-domain datasets, although the results vary with the data preprocessing. This is true even more so of re-ranking, which is much better when the BPE vocabulary is learned on all domains. Combining both shortlisting and re-ranking always brings in a slight improvement over just re-ranking on the out-of-domain datasets, but its effectiveness is again preprocessing-dependent. 

This clearly shows that the IBM model implemented by fast-align can learn information complementary to NMT, and interpolating them is beneficial for achieving higher BLEU scores in out-of-domain settings. The lower BLEU scores on in-domain test sets are related to the aggressive shortlist we used: by increasing the output layer limit from 10 to 50 most probable tokens, we achieve identical BLEU scores as the baseline.

\subsection{BLEU breakdown}

According to BLEU scores reported earlier, shortlisting and re-ranking are beneficial to NMT domain adaptation (in a relatively low-resource condition). We try to understand what contributes to the increase in BLEU scores by breaking down BLEU scores into n-gram precisions and length (brevity) penalty. In Table~\ref{table-bleu-scores} we list the numbers for \textit{law} and \textit{IT} domains, under the ``BPE trained on all'' setting in Table~\ref{results_tab}, on which we have seen the largest leap of BLEU scores. Additionally, we include METEOR which focuses on n-gram overlap and is not influenced by the output length. 

From the table, shortlisting always significantly boosts n-gram precisions, whereas re-ranking alone decreases them. On the other hand, re-ranking ``rectifies'' the output length, leading to a better brevity penalty (closer to 1) compared to baseline or shortlisting. When using both together, we see slight improvement on both n-gram accuracies and length penalty over baseline. This is expected: shortlisting provides extra word alignment information which aids lexical accuracy; re-ranking favours the hypothesis with an average length, since too long or too short hypotheses will receive a lower similarity score. This implies that shortlisting and re-ranking enhance BLEU from different aspects. 

On the contrary, according to METEOR, re-ranking is outperformed even by the baseline, leading to a negative METEOR change. This suggests that the bump in BLEU scores by re-ranking is only due to an improved brevity penalty. While it encourages a more desirable length, re-ranking does not produce better lexical choices. Adequacy wise, shortlisting is proven to be a more constructive method.

\section{Limitations}
\label{sec:limitations}

Although our methods show promising gains on the low-resource domains, we found that they have limited application when the domain mismatch is too great or there is sufficient resource available.

\begin{table*}[tbhp]
\centering
\begin{threeparttable}
\begin{tabular}{|l|rrrrr|}
\hline
Domain  & law & medical  & subtitles\tnote{\dag}  & IT & Koran \\
\hline
Number of sentences               & 695k  & 1M    & 1M   & 372k  & 529k  \\
\hline
Avg. original sentence length     & 22.1  & 12.5  & 8.0    & 7.5   & 20.4  \\
Avg. BPE sentence length          & 30.4  & 14.3  & 11.1  & 12.7  & 24.1  \\
\hline
Vocab size, appearing \textgreater 20 times & 34k & 36k   & 30k   & 15k  & 20k \\
Vocab overlap with \textit{medical}      & 11.5k & 36k   & 9.0k   & 5.8k  & 5.1k  \\
\hline
\end{tabular}
\begin{tablenotes}
\small \item[\dag]The \textit{subtitles} corpus was sampled down from 20M to 1M sentence pairs.
\end{tablenotes}
\end{threeparttable}
\caption{Corpus statistics for the different domains.}
\label{tab:domain-stats}
\end{table*}

\begin{table*}[tbhp]
\centering
\begin{tabular}{|l|l|}
    \hline
    German & English \\
    \hline
    \small sein Pilot hat nicht die volle Kontrolle .  & \small its p@@ il@@ ot is@@ n't in control . \\
    \small und Z@@ eth@@ rid ? nur einen Strei@@ f@@ sch@@ uss . & \small and , Z@@ eth@@ rid , just gr@@ aze it . \\
    \hline
\end{tabular}
\caption{Two random German-English sentence pairs from the \textit{subtitles} dataset after BPE.}
\label{tab:subtitles}
\end{table*}

\subsection{Large domain mismatch}

In order to better interpret why our methods are much more helpful on some domains than on others (e.g.\ \textit{law} versus \textit{Koran}), we gather statistics of our test sets in different domains in Table ~\ref{tab:domain-stats}, to reflect the distance between domains. We compute vocabulary overlap with the in-domain \textit{medical} data, for each cleaned corpus prior to BPE encoding. We count only words that are seen at least 20 times, and we sample 5\% of the \textit{subtitles} corpus, since it is orders of magnitude larger than the rest. After down-sampling, it has a similar size to the \textit{medical} corpora. We also compute the average sentence length before and after BPE to determine the extent to which BPE transforms the original sentences. Both vocabulary overlap and sentence lengths implicitly reflect the degree of domain mismatch.

For our most difficult datasets, \textit{subtitles} and \textit{Koran}, we see the lowest vocabulary overlap with the \textit{medical} domain of just 22-25\%. When BPE is applied to these corpora, the sentence length increases by 20-30\%. In practice, extensive and uneven BPE segmentation on named entities makes it difficult for the IBM model to produce interpretable and meaningful alignments to aid a translation model, as exemplified in Table ~\ref{tab:subtitles}. 

This means that lexical shortlisting is useful when we have a model trained on a relatively modest amount of data, and the out-of-domain dataset which we adapt to should share a reasonable amount of vocabulary. When the vocabulary overlap is too small, 
shortlisting drastically loses its effectiveness as a domain adaptation tool. 

We confirm this by experimenting with another domain-distant, and extremely low-resource Burmese-English scenario. The training data consists 18k sentence pairs from \textit{news} articles \citep{myen_2}, and the out-of-domain test set is a \textit{Bible} corpus \citep{100bible_cite}. The English side of the training set has 51k unique words, while the \textit{Bible} set has 29k. The vocabulary overlap between the training and test sets is just 6k words, equivalent to 21\% of the test vocabulary. After BPE, The average test sentence length increases from 25 to 38 tokens, seeing a 52\% rise. We adopt the model trained by \citet{aji-etal-2020-neural} and add shortlisting during decoding. The results displayed in Table~\ref{results_myen} show that out-of-domain performance does not improve with shortlisting, likely due to the high degree of vocabulary mismatch.

\begin{table}[ht!]
\centering
\begin{tabular}{|l|cc|}
\hline
                  & baseline & shortlist     \\
\hline
news (in-domain)  & 18.00    & 15.7            \\
Bible             & 0.2      & 0.2            \\
\hline
\end{tabular}
\caption{Very low-resource Burmese-English results.}
\label{results_myen}
\end{table}

\subsection{Availability of resources}
Our previous German-English experiments are carried out under an artificially resource-constrained condition. We therefore verify the potential of our methods in a high-resource setting too, by applying shortlisting on a WMT19 German-English submission from Microsoft \citep{junczysdowmunt:2019:WMT}, and evaluating on the same out-of-domain datasets. 

\begin{table}[!ht]
\centering
\begin{tabular}{|l|cc|c|}
\hline
 & \multicolumn{2}{c|}{Microsoft WMT19} & low-resource \\
          & baseline & shortlist & baseline  \\
\hline
medical   & 14.4 & 14.4 & 61.4 \\
Koran     & 0.0  & 0.0  & 0.8  \\
law       & 8.7  & 8.7  & 17.8 \\
IT        & 15.4 & 15.4 & 15.7 \\
subtitles & 1.0  & 1.0  & 2.6 \\
\hline
\end{tabular}

\caption{High-resource German-English results.} 
\label{results_ms}
\end{table}

Results in Table~\ref{results_ms} show that shortlisting has no impact on BLEU comparing to the baseline. We conclude that given a high-resource setting and the apparent large domain mismatch, the IBM model's alignments do not contribute additional information to the model. This finding is corroborated by \citet{li-etal-2019-word}: IBM model alignments are mostly better at capturing function words, not content words, compared to a neural model. Furthermore, unnatural and aggressive BPE segmentation on out-of-domain text (e.g.\ Table~\ref{tab:subtitles}) could result in a lexical shortlist not capturing any meaningful alignment.

We add our low-resource baseline trained on \textit{medical} with BPE on all data (Section~\ref{sec_results_and_analysis} Table~\ref{results_tab}) to comparison, and find that it surpasses the huge WMT model on each domain. Two reasons account for this: a WMT model is heavily biased to the news domain only; and the BPE scheme learned from news data is inferior to one learned from the out-of-domain datasets, when being evaluated on these domains.

\section{Conclusion}

We explore computationally cheap methods to improve neural machine translation performance in out-of-domain settings. We suggest that adding a lexical shortlist trained on the same data is always beneficial. While re-ranking also improves BLEU, it targets the BLEU brevity penalty, and does not produce better word choices. Although our results are promising in a low-resource condition, they do not transfer well to a scenario with very distant domains or sufficient resources. Our analysis shows that this is due to little vocabulary overlap, and the limited contribution from the IBM model under out-of-domain BPE segmentation.

\section*{Acknowledgements}

We would like to thank the anonymous reviewers, as well as the members of the AGORA research group for their valuable comments.

\lettrine[image=true, lines=2, findent=1ex, nindent=0ex, loversize=.15]{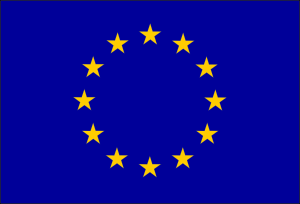}%
{T}his work was conducted within the scope of the \ProjectType\ \textit{\ProjectName}, which has received funding from the European Union's Horizon 2020 research and innovation programme under grant agreement No \GrantNo. Any communication or publication related to the action, made by the beneficiaries jointly or individually in any form and using any means, shall indicate that it reflects only the author's view and that the Agency is not responsible for any use that may be made of the information it contains.

This research is based upon work supported in part by the Office of the Director of National Intelligence (ODNI), Intelligence Advanced Research Projects Activity (IARPA), via contract \#FA8650-17-C-9117. The views and conclusions contained herein are those of the authors and should not be interpreted as necessarily representing the official policies, either expressed or implied, of ODNI, IARPA, or the U.S. Government. The U.S. Government is authorized to reproduce and distribute reprints for governmental purposes notwithstanding any copyright annotation therein.

\bibliography{emnlp2020}
\bibliographystyle{acl_natbib}




\end{document}